\documentclass{article}

\usepackage{nips15submit_e,times}
\usepackage{algorithm}
\usepackage{algorithmicx}
\usepackage{algpseudocode}
\usepackage{hyperref}
\usepackage{url}
\usepackage{amsmath}
\usepackage{amssymb}
\usepackage{textcomp}
\usepackage{nth}
\usepackage{todonotes}
\usepackage{graphicx}
\usepackage{booktabs}
\usepackage[justification=centering]{caption}
\usepackage{bbm}

\presetkeys{todonotes}{inline, fancyline}{}
\usepackage[marginparwidth=3cm, marginparsep=0.3cm, outer=3.6cm]{geometry}

\usepackage{array}
\newcolumntype{L}[1]{>{\raggedright\let\newline\\\arraybackslash\hspace{0pt}}m{#1}}
\newcolumntype{C}[1]{>{\centering\let\newline\\\arraybackslash\hspace{0pt}}m{#1}}
\newcolumntype{R}[1]{>{\raggedleft\let\newline\\\arraybackslash\hspace{0pt}}m{#1}}

\title{Large-scale Validation of Counterfactual \\Learning Methods: A Test-Bed}

\author{
Damien Lefortier\thanks{This work was done while working at Criteo.} \\
Facebook \& University of
Amsterdam \\
dlefortier@fb.com\\
\And
Adith Swaminathan \\
Cornell University, Ithaca, NY \\
adith@cs.cornell.edu\\
\And
Xiaotao Gu\\
Tsinghua University, Beijing, China\\
gxt13@mails.tsinghua.edu.cn\\
\And
Thorsten Joachims\\
Cornell University, Ithaca, NY \\
tj@cs.cornell.edu\\
\And
Maarten de Rijke\\
University of Amsterdam \\
derijke@uva.nl\\
}

\nipsfinalcopy 

\begin{document}

\maketitle

\begin{abstract}
The ability to perform effective off-policy learning would revolutionize the process of building better interactive systems, such as search engines and recommendation systems for e-commerce,
computational advertising and news. Recent approaches for off-policy evaluation and learning in these settings appear promising \cite{bottou2013counterfactual, swaminathan2015batch}. With this paper, we provide real-world data and a standardized test-bed to systematically investigate these algorithms  using data from display advertising. In particular, we consider the problem of filling a banner ad with an aggregate of multiple products the user may want to purchase. This paper presents our test-bed, the sanity checks we ran to ensure its validity, and shows results comparing 
state-of-the-art off-policy learning methods
like doubly robust optimization \cite{dudik2011},
POEM  \cite{swaminathan2015batch}, and reductions to supervised learning using regression baselines.
Our results show experimental evidence that recent off-policy learning methods can improve upon state-of-the-art supervised learning techniques on a large-scale real-world data set.
\end{abstract}

\section{Introduction}
Effective learning methods for optimizing policies based on logged user-interaction data have the potential to revolutionize the process of building better interactive systems. Unlike the industry standard of using expert judgments for training, such learning methods could directly optimize user-centric performance measures, they would not require interactive experimental control like online algorithms, and they would not be subject to the data bottlenecks and latency inherent in A/B testing.

Recent approaches for off-policy evaluation and learning in these settings appear promising \cite{bottou2013counterfactual, swaminathan2015batch, swaminathan2015self}, but highlight the need for accurately logging propensities of the logged actions.  
With this paper, we provide the first public dataset that contains accurately logged propensities for the problem of Batch Learning from Bandit Feedback (BLBF). We use data from Criteo, a leader in the display advertising space. In addition to providing the data, we propose an evaluation methodology for running BLBF learning experiments and a standardized test-bed that allows the research community to systematically investigate BLBF algorithms. 

At a high level, a BLBF algorithm operates in the contextual bandit setting and solves the following learning task:
\begin{enumerate}
\item Take as input: $\{ \pi_0, \langle x_i, y_i, \delta_i \rangle_{i=1}^n \}$. $\pi_0$ encodes the system from which the logs were collected,
$x$ denotes the input to the system, $y$ denotes the output predicted by the system and $\delta$ is a number encoding the observed online metric for the output that was predicted; 
\item Produce as output: $\pi$, a new policy that maps $x \mapsto y$; and
\item Such that $\pi$ will perform well (according to the metric $\delta$) \emph{if it were} deployed online.
\end{enumerate}

We elaborate on the definitions of $x, y, \delta, \pi_0$ as logged in our dataset in the next section. Since past research on BLBF was limited due to the availability of an appropriate dataset, we hope that our test-bed will spur research on several aspects of BLBF and off-policy evaluation, including the following:
\begin{enumerate}
\item New training objectives, learning algorithms, and regularization mechanisms for BLBF;
\item Improved model selection procedures (analogous to cross-validation for supervised learning);
\item Effective and tractable policy classes $\pi \in \Pi$ for the specified task $x \mapsto y$; and
\item Algorithms that can scale to massive amounts of data.
\end{enumerate}

The rest of this paper is organized as follows. In Section \ref{sec:challenge}, we describe our standardized test-bed for the evaluation of off-policy learning methods. Then, in Section~\ref{sec:checks}, we describe a set of sanity checks that we used on our dataset to ensure its validity and that can be applied generally when gathering data for off-policy learning and evaluation. Finally, in Section~\ref{sec:results}, we show results comparing state-of-the-art off-policy learning methods
like doubly robust optimization \cite{dudik2011},
POEM  \cite{swaminathan2015batch}, and reductions to supervised learning using regression baselines. Our results show, for the first time, experimental evidence that recent off-policy learning methods can improve upon state-of-the-art supervised learning techniques on a large-scale real-world data set.

\section{Dataset}
\label{sec:challenge}
\begin{figure}[ht]
    \centering
    \includegraphics[scale=0.4]{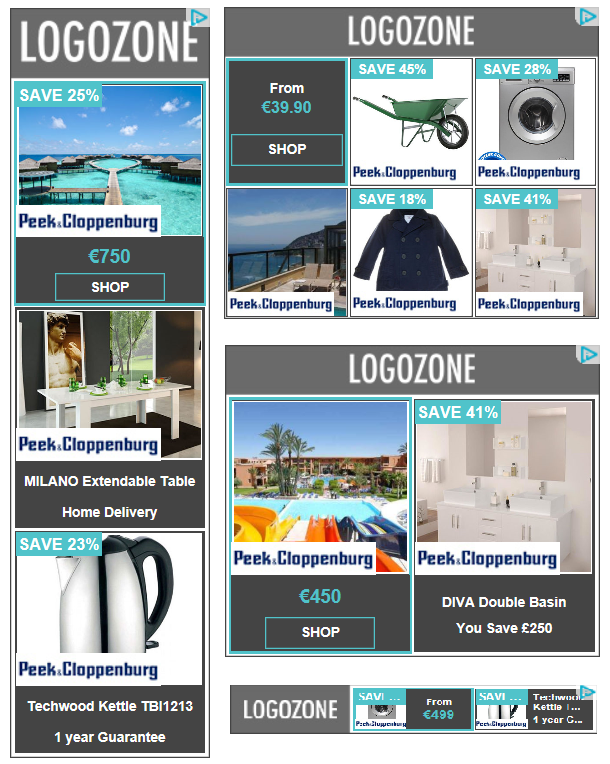}
    \caption{Four examples of ads used in display advertising: a vertical ad, a grid, and two horizontal ads (mock-ups).}
    \label{BannerExamples}
\end{figure}

We create our test-bed using data from display advertising, similar to the Kaggle challenge hosted by Criteo in 2014 to compare CTR prediction algorithms.\footnote{\url{https://www.kaggle.com/c/criteo-display-ad-challenge}} However, in this paper, we do not aim to build clickthrough or conversion prediction models for bidding in real-time auctions \cite{chapelle2014simple, vasile2016cost}. Instead, we consider the problem of filling a banner ad with an aggregate of multiple products the user may want to purchase. This part of the system takes place after the bidding agent has won the auction. In this context, each ad has one of many banner types, which differ in the number of products they contain and in their layout as shown in Figure~\ref{BannerExamples}. The task is to choose the products to display in the ad knowing the banner type in order to maximize the number of clicks. This task is thus very different from the Kaggle challenge.

In this setting of choosing the best products to fill the banner ad, we can easily gather exploration data where the placement of the products in the banner ad is randomized, without incurring a prohibitive cost unlike in Web search for which such exploration is much more costly (see, e.g., \cite{vorobev2015gathering, lefortier2014online}). Our logging policy uses randomization aggressively, while being very different from a uniformly random policy.

Each banner type corresponds to a different \emph{look \& feel} of the banner ad. Banner ads can differ in the number of products, size, geometry (vertical, horizontal, \ldots), background color and in the data shown (with or without a product description or a call to action); these we call the \emph{fixed} attributes. Banner types may also have \emph{dynamic aspects} such as some form of pagination (multiple pages of products) or an animation. Some examples are shown in Figure~\ref{BannerExamples}.
Throughout the paper, we label positions in each banner type from 1 to $N$ from left to right and from top to bottom. Thus 1 is the top left position. 

For each user impression, we denote a user context by $c$, the number of slots in the banner type by $l_c$, and the candidate pool of products $p$ by $P_c$.
Each context $c$ and product $p$ pair is described by features $\phi(c,p)$. The input $x$ to the system encodes $c, P_c, \{ \phi(c,p) : p \in P_c\}$.
The logging policy $\pi_0$
stochastically selects products to construct a banner by first computing non-negative scores $f_p$ for all candidate products $p \in P_c$, and 
using a Plackett-Luce ranking model (i.e., sampling without replacement from the multinomial distribution defined by the $f_p$  scores):
\begin{eqnarray}
P(slot1 = p) = \frac{f_p}{\sum_{\{p' \in P_c\}}f_{p'}} \qquad P(slot2 = p'\mid slot1=p) = \frac{f_{p'}}{\sum_{\{p^{\dagger} \in P_c \wedge p^{\dagger} \ne p\}}f_{p^{\dagger}}}, \quad \text{etc.}
\end{eqnarray}
 The propensity of a chosen banner ad $\langle p_1, p_2, \dots \rangle$ is $P(slot1 = p_1)\cdot P(slot2 = p_2 \mid slot1=p_1)\cdot \ldots$.
 With these propensities in hand, we can counterfactually evaluate any banner-filling policy in an unbiased way using inverse propensity scoring \cite{Li2011}.

The following was logged, committing to a single feature encoding $\phi(c, p)$ and a single $\pi_0$ that produces the scores $f$ for the entire duration of data collection. 
\begin{itemize}
\item Record the feature vector $\phi(c, p)$ for all products in the candidate set $P_c$;
\item Record the selected products sampled from $\pi_0$ via the Plackett-Luce model and its propensity;
\item Record the click/no click and their location(s) in the banner.
\end{itemize}
The format of this data is:

\texttt{ \tiny \noindent
example \$\{exID\}: \$\{hashID\} \$\{wasAdClicked\} \$\{propensity\} \$\{nbSlots\} \$\{nbCandidates\} \$\{displayFeat1\}:\$\{v\_1\} \dots
}

\texttt{ \tiny \noindent
\$\{wasProduct1Clicked\} exid:\$\{exID\} \$\{productFeat1\_1\}:\$\{v1\_1\} ...
}

\texttt{ \tiny \noindent
...
}

\texttt{ \tiny \noindent
\$\{wasProductMClicked\} exid:\$\{exID\} \$\{productFeatM\_1\}:\$\{vM\_1\} ...
}

Each impression is represented by $M+1$ lines where $M$ is the cardinality of $P_c$ and the first line is a header containing summary information.
Note that the first \$\{nbSlots\} candidates correspond to the displayed products ordered by position (consequently, \$\{wasProductMClicked\}
information for all other candidates is irrelevant). There are $35$ features. Display features are context features or banner type features, which are constant for all candidate products in a given impression. Each unique quadruplet of feature IDs $\langle 1, 2, 3, 5 \rangle$ correspond to a unique banner type. Product features are based on the similarity and/or complementarity of the candidate products with historical products seen by the user on the advertiser's website. We also included interaction terms between some of these features directly in the dataset to limit the amount of feature engineering required to get a good policy. 
Features 1 and 2 are numerical, while all other features are categorical. Some categorical features are multi-valued, which means that they can take more than one value for the same product (order does not matter). 
Note that the example ID is increasing with time, allowing temporal slices for evaluation \cite{mcmahan2013ad}, although we do not enforce this for our test-bed.
Importantly, non-clicked examples were sub-sampled aggressively to reduce the dataset size and \emph{we kept only a random $10\%$ sub-sample of them}. So, one needs to account for this during learning and evaluation -- the evaluator we provide with the test-bed accounts for this sub-sampling.
 
The result is a dataset of over $103$ million ad impressions. In this dataset, we have:
\begin{itemize}
    \item $8500+$ banner types with the top $10$ banner types representing $30\%$ of the total number of ad impressions, the top $50$ about $65\%$, and the top $100$ about $80\%$.
    \item The number of displayed products is between $1$ and $6$ included.
    \item There are over $21M$ impressions for $1$-slot banners, over $35M$ for $2$-slot, almost $23M$ for $3$-slot, $7M$ for $4$-slot, $3M$ for $5$-slot and over $14M$ for $6$-slot banners.
    \item  The size of the candidate pool $P_c$ is about $10$ times (upper bound) larger than the number of products to display in the ad.
\end{itemize}
This dataset is hosted on Amazon AWS (35GB gzipped / 256GB raw).
Details for accessing and processing the data are available at \url{http://www.cs.cornell.edu/~adith/Criteo/}. 

\section{Sanity Checks}
\label{sec:checks}
The work-horse of counterfactual evaluation is Inverse Propensity Scoring (IPS) \cite{Rosenbaum1983,Li2011}. IPS requires accurate propensities, and, to a crude approximation, produces estimates with variance that scales roughly with the range of the inverse propensities. In Table~\ref{tab:minmaxmean_propensity}, we report the number of impressions and the average and largest inverse propensities, partitioned by \$\{nbSlots\}. When constructing confidence intervals for importance weighted estimates like IPS, we often appeal to asymptotic normality of large sample averages \cite{Owen2013}. However,
if the inverse propensities are very large relative to the number of samples (as we can see for $\$\{nbSlots\} \ge 4$), the asymptotic normality assumption will probably be violated.

\begin{table}[ht]
	\caption{Number of impressions and propensity statistics computed for slices of traffic with $k$-slot banners, $1 \le k \le 6$. Estimated sample size ($\hat{N}$) corrects for $10\%$ sub-sampling of unclicked impressions.}
	\label{tab:minmaxmean_propensity}
	\begin{center}
		\begin{tabular}{|c|c|c|c|c|c|c|}
			\toprule
			\textbf{\#Slots} & \textbf{1} & \textbf{2} & \textbf{3} & \textbf{4} & \textbf{5} & \textbf{6}\\
			\midrule
			\#Impressions & $2.13e+07$ & $3.55e\!+\!07$ & $2.27e\!+\!07$ & $6.92e\!+\!06$ & $2.95e\!+\!06$ & $1.40e\!+\!07$ \\
			$\hat{N}$ & $2.03e+08$ & $3.39e\!+\!08$ & $2.15e\!+\!08$ & $6.14e\!+\!07$ & $2.65e\!+\!07$ & $1.30e\!+\!08$ \\
			Avg(InvPropensity) & $11.96$ & $3.29e\!+\!02$ &  $1.87e\!+\!04$ & $2.29e\!+\!06$ & $2.62e\!+\!07$ & $3.51e\!+\!09$\\
            Max(InvPropensity) & $5.36e\!+\!05$ & $3.38e\!+\!08$ &  $3.23e\!+\!10$ & $9.78e\!+\!12$ & $2.03e\!+\!12$ & $2.34e\!+\!15$\\
 			\bottomrule
		\end{tabular}
	\end{center}
\end{table}

There are some simple statistical tests that can be run to detect some issues with inaccurately logged propensities \cite{li2015counterfactual}.
These \emph{arithmetic} and \emph{harmonic} tests, however, require that the candidate actions available for each impression are fixed a priori.
In our scenario, we have a context-dependent candidate set that precludes running these tests, so we propose a more general class of diagnostics that can detect some systematic biases and issues in propensity-logged datasets.

Some notation: $x_i \overset{iid}{\sim} \Pr(X); \qquad y_i \sim \pi_0(Y \mid x_i); \qquad \delta_i \sim \Pr(\Delta \mid x_i, y_i)$. The propensity for the logging policy $\pi_0$ to take the logged action $y_i$ in context $x_i$ is denoted $q_i \equiv \pi_0(y_i \mid x_i)$.
If the propensities are correctly logged, then the expected importance weight should be $1$ for any new banner-filling policy $\pi(Y \mid x)$. 
Formally, we have the following:
\begin{equation}
\hat{C}(\pi) = \frac{1}{N}\sum_{i=1}^N \frac{\pi(y_i \mid x_i)}{q_i}  \simeq 1.
\end{equation}
The IPS estimate for a new policy is simply:
\begin{equation}
\hat{R}(\pi) = \frac{1}{N}\sum_{i=1}^N \delta_i \frac{\pi(y_i \mid x_i)}{q_i}.
\end{equation}
These equations are valid when $\pi_0$ has full support, as our logging system does: $\pi_0(y \mid x) > 0 \qquad \forall x, y$. 
The self-normalized estimator \cite{Hesterberg1995,swaminathan2015self} is: 
\begin{equation}
\hat{R}_{snips}(\pi) = \frac{\hat{R}(\pi) }{ \hat{C}(\pi)} .    
\end{equation}

Remember that we sub-sampled non-clicked impressions. Sub-sampling is indicated by the binary random variable $o_i$:
\begin{equation}
    o_i \sim \Pr(O = 1 \mid \delta) = 
        \begin{cases}
        0.1 & \text{if } \delta = 0, \\
        1 & \text{otherwise. }
        \end{cases}
\end{equation}
The IPS estimate and the diagnostic above are not computable in our case since they require all data-points before sub-sampling. So, we use the following straightforward modification to use only our $N$ sub-sampled data-points instead.

First, we estimate the number of data-points before sub-sampling $\hat{N}$ only using samples where $ o_i = 1$:
\begin{equation}
\hat{N} = \sum_{i=1}^N \frac{\mathbbm{1}\{ o_i = 1\}}{\Pr(O=1\mid \delta_i)} = \#\{\delta = 1\} + 10 \#\{\delta = 0\}.
\end{equation}
$\hat{N}$ is an unbiased estimate of $N = \sum_{i=1}^N 1$ since $\mathbb{E}_{(x_i, y_i, \delta_i)} \mathbb{E}_{o_i \sim \Pr(O\mid \delta_i)} \left[ \frac{\mathbbm{1}\{ o_i = 1\}}{\Pr(O=1\mid \delta_i)} \right] = \mathbb{E}_{(x_i, y_i, \delta_i)} 1 = 1$.
Next, consider estimating $R(\pi) = \mathbb{E}_{(x_i, y_i, \delta_i)} \delta_i \frac{\pi(y_i \mid x_i)}{q_i}$ as:
\begin{equation}
    \hat{R}(\pi) = \frac{1}{\hat{N}} \sum_{i=1}^N \delta_i \frac{\pi(y_i \mid x_i)}{q_i} \frac{\mathbbm{1}\{ o_i = 1\}}{\Pr(O=1\mid \delta_i)}.
\end{equation}
Again, $\mathbb{E}_{(x_i, y_i, \delta_i)} \mathbb{E}_{o_i \sim \Pr(O\mid \delta_i)} \left[ \delta_i \frac{\pi(y_i \mid x_i)}{q_i} \frac{\mathbbm{1}\{ o_i = 1\}}{\Pr(O=1\mid \delta_i)} \right] = \mathbb{E}_{(x_i, y_i, \delta_i)} \delta_i \frac{\pi(y_i \mid x_i)}{q_i}$.
Hence, the sum in the numerator of $\hat{R}(\pi)$ is, in expectation, $N R(\pi)$, while the normalizing constant $\hat{N}$ is, in expectation, $N$.
Ratios of expectations are not equal to the expectation of a ratio, so we expect a small bias in this estimate but it is easy to show that this estimate is asymptotically consistent. 

Finally consider estimating $C(\pi) = \mathbb{E}_{(x_i, y_i)} \frac{\pi(y_i \mid x_i)}{q_i} = 1$ as:
\begin{equation}
    \hat{C}(\pi) = \frac{1}{\hat{N}} \sum_{i=1}^N \frac{\pi(y_i \mid x_i)}{q_i} \frac{\mathbbm{1}\{ o_i = 1\}}{\Pr(O=1\mid \delta_i)}.
\end{equation}
Again, $\mathbb{E}_{(x_i, y_i, \delta_i)} \mathbb{E}_{o_i \sim \Pr(O\mid \delta_i)} \left[ \frac{\pi(y_i \mid x_i)}{q_i} \frac{\mathbbm{1}\{ o_i = 1\}}{\Pr(O=1\mid \delta_i)} \right] = \mathbb{E}_{(x_i, y_i, \delta_i)} \frac{\pi(y_i \mid x_i)}{q_i} = 1$. The sum in the numerator of $\hat{C}(\pi)$ is, in expectation, $N$ as is the denominator. Again, we expect this estimate to have a small bias but to remain asymptotically consistent.
The computable variant of the self-normalized IPS estimator simply uses the computable $\hat{R}(\pi)$ and $\hat{C}(\pi)$ in its definition: 
$\hat{R}_{snips}(\pi) = \hat{R}(\pi) / \hat{C}(\pi)$. 

We use a family of new policies $\pi_\epsilon$, parametrized by $0 \le \epsilon \le 1$ to diagnose $\hat{C}(\pi)$ and the expected behavior of IPS estimates $\hat{R}(\pi)$.
The policy $\pi_\epsilon$ behaves like a uniformly random ranking policy with probability $\epsilon$, and with probability $1 - \epsilon$, behaves like the logging policy.
Formally, for an impression with context $x_i$, $\left|\mathcal{Y}\right|$ possible actions (e.g., rankings of candidate products), and logged action $y_i$, the probability for choosing $y_i$ under the new policy $\pi_\epsilon$ is:
\begin{equation}
\pi_\epsilon(y_i \mid x_i) = \epsilon \frac{1}{\left|\mathcal{Y}\right|} + (1 - \epsilon) \pi_0(y_i \mid x_i).
\end{equation}

As we vary $\epsilon$ away from $0$, the new policy looks more different than the logging policy $\pi_0$ on the logged impressions.
In Tables~\ref{tab:epsilonlogger1},\ref{tab:epsilonlogger2},\ref{tab:epsilonlogger3} we report $\hat{C}(\pi_\epsilon)$ and a $99\%$ confidence interval assuming asymptotic normality, for different choices of $\epsilon$. We also report the IPS-estimated clickthrough rates for these policies $\hat{R}(\pi_\epsilon)$, their standard error ($99\%$ CI), and finally, their self-normalized IPS-estimates \cite{Hesterberg1995,swaminathan2015self}.

\begin{table}[htb]
	\caption{Diagnostics and IPS-estimated clickthrough rates for different policies $\pi_\epsilon$ evaluated on slices of traffic with $k$-slot banners, $k \in \{1, 2 \}$. $\epsilon$ interpolates between the logging policy ($\epsilon = 0$) and the uniform random policy ($\epsilon = 1$). Error bars are $99\%$ confidence intervals under a normal distribution.}
	\label{tab:epsilonlogger1}
	\begin{center}
		\begin{tabular}{|c|ccc|ccc|}
			\toprule
			\textbf{\#Slots} & 
			\multicolumn{3}{|c|}{\textbf{1}}  &
			\multicolumn{3}{|c|}{\textbf{2}}
			\\
			\midrule
			$\epsilon$ & $\hat{C}(\pi_\epsilon)$ & $\hat{R}(\pi_\epsilon) \times 10^4$ & $\frac{\hat{R}(\pi_\epsilon) \times 10^4 }{ \hat{C}(\pi_\epsilon)}$ & $\hat{C}(\pi_\epsilon)$ & $\hat{R}(\pi_\epsilon) \times 10^4$ & $\frac{\hat{R}(\pi_\epsilon) \times 10^4}{ \hat{C}(\pi_\epsilon)}$ \\
\midrule
$0$ & $1.000\! \pm \!0.000$ & $53.604\! \pm \!0.129$ & $53.604$ & $1.000\! \pm \!0.000$ & $52.554\! \pm \!0.099$ & $52.554$ \\
$2^{-10}$ & $1.000\! \pm \!0.000$ & $53.598\! \pm \!0.129$ & $53.599$ & $1.000\! \pm \!0.000$ & $52.541\! \pm \!0.099$ & $52.545$ \\
$2^{-9}$ & $1.000\! \pm \!0.000$ & $53.593\! \pm \!0.130$ & $53.595$ & $1.000\! \pm \!0.000$ & $52.529\! \pm \!0.101$ & $52.536$ \\
$2^{-8}$ & $1.000\! \pm \!0.000$ & $53.582\! \pm \!0.131$ & $53.585$ & $1.000\! \pm \!0.000$ & $52.503\! \pm \!0.107$ & $52.517$ \\
$2^{-7}$ & $1.000\! \pm \!0.000$ & $53.560\! \pm \!0.138$ & $53.567$ & $0.999\! \pm \!0.000$ & $52.453\! \pm \!0.129$ & $52.481$ \\
$2^{-6}$ & $1.000\! \pm \!0.000$ & $53.516\! \pm \!0.163$ & $53.531$ & $0.999\! \pm \!0.001$ & $52.351\! \pm \!0.193$ & $52.407$ \\
$2^{-5}$ & $0.999\! \pm \!0.000$ & $53.428\! \pm \!0.236$ & $53.457$ & $0.998\! \pm \!0.002$ & $52.148\! \pm \!0.346$ & $52.260$ \\
$2^{-4}$ & $0.999\! \pm \!0.001$ & $53.251\! \pm \!0.416$ & $53.311$ & $0.996\! \pm \!0.003$ & $51.742\! \pm \!0.671$ & $51.965$ \\
$2^{-3}$ & $0.998\! \pm \!0.001$ & $52.899\! \pm \!0.802$ & $53.017$ & $0.991\! \pm \!0.006$ & $50.929\! \pm \!1.331$ & $51.370$ \\
$2^{-2}$ & $0.996\! \pm \!0.003$ & $52.194\! \pm \!1.589$ & $52.428$ & $0.983\! \pm \!0.012$ & $49.305\! \pm \!2.657$ & $50.166$ \\
$2^{-1}$ & $0.991\! \pm \!0.006$ & $50.785\! \pm \!3.171$ & $51.241$ & $0.966\! \pm \!0.024$ & $46.056\! \pm \!5.312$ & $47.693$ \\
$1$ & $0.982\! \pm \!0.012$ & $47.966\! \pm \!6.338$ & $48.836$ & $0.931\! \pm \!0.048$ & $39.557\! \pm \!10.623$ & $42.473$ \\
\bottomrule

		\end{tabular}
	\end{center}
\end{table}

\begin{table}[htb]
	\caption{Diagnostics for different policies $\pi_\epsilon$ evaluated on slices of traffic with $k$-slot banners, $k \in \{3, 4 \}$. 
	Error bars are $99\%$ confidence intervals under a normal distribution.}
	\label{tab:epsilonlogger2}
	\begin{center}
		\begin{tabular}{|c|ccc|ccc|}
			\toprule
			\textbf{\#Slots} & 
			\multicolumn{3}{|c|}{\textbf{3}}  &
			\multicolumn{3}{|c|}{\textbf{4}} \\
			\midrule
			$\epsilon$ & $\hat{C}(\pi_\epsilon)$ & $\hat{R}(\pi_\epsilon) \times 10^4$ & $\frac{\hat{R}(\pi_\epsilon) \times 10^4 }{ \hat{C}(\pi_\epsilon)}$ & $\hat{C}(\pi_\epsilon)$ & $\hat{R}(\pi_\epsilon) \times 10^4$ & $\frac{\hat{R}(\pi_\epsilon) \times 10^4}{ \hat{C}(\pi_\epsilon)}$ \\
\midrule
$0$ & $1.000\! \pm \!0.000$ & $64.298\! \pm \!0.137$ & $64.298$ & $1.000\! \pm \!0.000$ & $141.114\! \pm \!0.366$ & $141.114$ \\
$2^{-10}$ & $1.000\! \pm \!0.000$ & $64.296\! \pm \!0.148$ & $64.305$ & $1.000\! \pm \!0.001$ & $141.065\! \pm \!0.366$ & $141.082$ \\
$2^{-9}$ & $1.000\! \pm \!0.000$ & $64.294\! \pm \!0.179$ & $64.312$ & $1.000\! \pm \!0.001$ & $141.015\! \pm \!0.366$ & $141.049$ \\
$2^{-8}$ & $0.999\! \pm \!0.000$ & $64.291\! \pm \!0.268$ & $64.326$ & $1.000\! \pm \!0.002$ & $140.916\! \pm \!0.368$ & $140.984$ \\
$2^{-7}$ & $0.999\! \pm \!0.001$ & $64.284\! \pm \!0.480$ & $64.354$ & $0.999\! \pm \!0.003$ & $140.717\! \pm \!0.378$ & $140.853$ \\
$2^{-6}$ & $0.998\! \pm \!0.001$ & $64.269\! \pm \!0.930$ & $64.410$ & $0.998\! \pm \!0.006$ & $140.320\! \pm \!0.413$ & $140.590$ \\
$2^{-5}$ & $0.996\! \pm \!0.003$ & $64.240\! \pm \!1.844$ & $64.523$ & $0.996\! \pm \!0.012$ & $139.526\! \pm \!0.534$ & $140.065$ \\
$2^{-4}$ & $0.991\! \pm \!0.006$ & $64.182\! \pm \!3.681$ & $64.750$ & $0.992\! \pm \!0.024$ & $137.937\! \pm \!0.863$ & $139.007$ \\
$2^{-3}$ & $0.982\! \pm \!0.011$ & $64.066\! \pm \!7.359$ & $65.211$ & $0.985\! \pm \!0.049$ & $134.761\! \pm \!1.610$ & $136.867$ \\
$2^{-2}$ & $0.965\! \pm \!0.023$ & $63.834\! \pm \!14.716$ & $66.157$ & $0.969\! \pm \!0.097$ & $128.407\! \pm \!3.161$ & $132.484$ \\
$2^{-1}$ & $0.930\! \pm \!0.045$ & $63.370\! \pm \!29.430$ & $68.156$ & $0.938\! \pm \!0.194$ & $115.700\! \pm \!6.295$ & $123.288$ \\
$1$ & $0.860\! \pm \!0.090$ & $62.443\! \pm \!58.860$ & $72.643$ & $0.877\! \pm \!0.389$ & $90.285\! \pm \!12.577$ & $102.960$ \\
\bottomrule

		\end{tabular}
	\end{center}
\end{table}

\begin{table}[htb]
	\caption{Diagnostics for different policies $\pi_\epsilon$ evaluated on slices of traffic with $k$-slot banners, $k \in \{5, 6 \}$. 
	Error bars are $99\%$ confidence intervals under a normal distribution.}
	\label{tab:epsilonlogger3}
	\begin{center}
		\begin{tabular}{|c|ccc|ccc|}
			\toprule
			\textbf{\#Slots} & 
			\multicolumn{3}{|c|}{\textbf{5}}  &
			\multicolumn{3}{|c|}{\textbf{6}} \\
			\midrule
			
			$\epsilon$ & $\hat{C}(\pi_\epsilon)$ & $\hat{R}(\pi_\epsilon) \times 10^4$ & $\frac{\hat{R}(\pi_\epsilon) \times 10^4 }{ \hat{C}(\pi_\epsilon)}$ & $\hat{C}(\pi_\epsilon)$ & $\hat{R}(\pi_\epsilon) \times 10^4$ & $\frac{\hat{R}(\pi_\epsilon) \times 10^4}{ \hat{C}(\pi_\epsilon)}$ \\
\midrule
$0$ & $1.000\! \pm \!0.000$ & $125.965\! \pm \!0.530$ & $125.965$ & $1.000\! \pm \!0.000$ & $90.620\! \pm \!0.206$ & $90.620$ \\
$2^{-10}$ & $0.999\! \pm \!0.000$ & $125.899\! \pm \!0.532$ & $125.976$ & $1.000\! \pm \!0.000$ & $90.579\! \pm \!0.207$ & $90.622$ \\
$2^{-9}$ & $0.999\! \pm \!0.001$ & $125.833\! \pm \!0.538$ & $125.988$ & $0.999\! \pm \!0.000$ & $90.537\! \pm \!0.210$ & $90.625$ \\
$2^{-8}$ & $0.998\! \pm \!0.001$ & $125.702\! \pm \!0.563$ & $126.011$ & $0.998\! \pm \!0.000$ & $90.454\! \pm \!0.222$ & $90.629$ \\
$2^{-7}$ & $0.995\! \pm \!0.001$ & $125.439\! \pm \!0.653$ & $126.057$ & $0.996\! \pm \!0.001$ & $90.289\! \pm \!0.264$ & $90.638$ \\
$2^{-6}$ & $0.990\! \pm \!0.002$ & $124.913\! \pm \!0.931$ & $126.149$ & $0.992\! \pm \!0.001$ & $89.957\! \pm \!0.389$ & $90.657$ \\
$2^{-5}$ & $0.980\! \pm \!0.004$ & $123.861\! \pm \!1.624$ & $126.337$ & $0.985\! \pm \!0.002$ & $89.293\! \pm \!0.691$ & $90.694$ \\
$2^{-4}$ & $0.961\! \pm \!0.007$ & $121.756\! \pm \!3.119$ & $126.725$ & $0.969\! \pm \!0.004$ & $87.967\! \pm \!1.336$ & $90.769$ \\
$2^{-3}$ & $0.922\! \pm \!0.014$ & $117.548\! \pm \!6.172$ & $127.549$ & $0.938\! \pm \!0.008$ & $85.313\! \pm \!2.649$ & $90.928$ \\
$2^{-2}$ & $0.843\! \pm \!0.029$ & $109.131\! \pm \!12.314$ & $129.428$ & $0.877\! \pm \!0.017$ & $80.006\! \pm \!5.287$ & $91.279$ \\
$2^{-1}$ & $0.686\! \pm \!0.057$ & $92.298\! \pm \!24.613$ & $134.475$ & $0.753\! \pm \!0.033$ & $69.392\! \pm \!10.568$ & $92.154$ \\
$1$ & $0.373\! \pm \!0.115$ & $58.631\! \pm \!49.221$ & $157.307$ & $0.506\! \pm \!0.066$ & $48.164\! \pm \!21.135$ & $95.185$ \\
\bottomrule

		\end{tabular}
	\end{center}
\end{table}

As we pick policies that differ from the logging policy, we see that 
the estimated variance of the IPS estimates (as reflected in their approximate $99\%$ confidence intervals) increases. Moreover, the control variate $\hat{C}(\pi_\epsilon)$ is systematically under-estimated.
This should caution us to not rely on a single point-estimate (e.g. only IPS or SNIPS). SNIPS can often provide a better bias-variance trade-off in these estimates, but can fail catastrophically when the variance is very high due to systematic under-estimation of $\hat{C}(\pi)$. Moreover, in these very high-variance situations (e.g. when $k \ge 3$ and $\epsilon \ge 2^{-2}$), the constructed confidence intervals are not reliable --- $C(\pi_\epsilon)$ clearly does not lie in the computed intervals.
Based on these sanity checks, we focus the evaluation set-up in Section~\ref{sec:results} on the $1$-slot case.

\section{Benchmarking Learning Algorithms}
\label{sec:results}
\subsection{Evaluation}
Estimates based on importance sampling have considerable variance when the number of slots increases.
We would thus need
tens of millions of impressions to estimate the CTR of slot-filling policies with high precision. To limit the risks of
people ``over-fitting to the variance'' by querying far away from our logging policy, we propose the following estimates for any policy:
\begin{itemize}
\item Report the inverse propensity scoring (IPS) \cite{Li2011} $\hat{R}(\pi)$ as well as the self-normalized (SN) estimate \cite{swaminathan2015self} for the new policy $\hat{R}(\pi) / \hat{C}(\pi)$ (self-normalized, so that
learnt policies cannot cheat by not having their importance weights sum to 1);
\item Compute the standard error of the IPS estimate (appealing to asymptotic normality), and report this error as an ``approximate confidence interval''.
\end{itemize}
This is provided in our evaluation software alongside the dataset \href{http://www.cs.cornell.edu/~adith/Criteo/}{online}. In this way, learning algorithms must
reason about bias/variance explicitly to reliably achieve better estimated CTR.
\subsection{Methods}
Consider a $1$-slot banner filling task defined using our dataset.
This $21M$ slice of traffic can be modeled as a logged contextual bandit problem with a small number of arms. 
This slice is further randomly divided into a $33-33-33\%$ train-validate-test split.
The following methods are benchmarked in the code accompanying this dataset release.
All these methods use a linear policy class $\pi \in \Pi_{lin}$ to map $x \mapsto y$ (i.e., score candidates using a linear scorer $w \cdot \phi(c, p)$), but differ in their training objectives. Their hyper-parameters are chosen to maximize $\hat{R}(\pi)$ on the validation set and their test-set estimates are reported in Table~\ref{tab:learnres}.

\begin{enumerate}
\item \textbf{Random}: A policy that picks $p \in P_c$ uniformly at random to display.
\item \textbf{Regression}: A reduction to supervised learning that predicts $\delta$ for every candidate action. The number of training epochs (ranging from $1 \dots 40$), regularization for Lasso (ranging from $10^{-8} \dots 10^{-4}$), and learning rate for SGD ($0.1, 1, 10$) are the hyper-parameters.
\item \textbf{IPS}: Directly optimizes $\hat{R}(\pi)$ evaluated on the \emph{training} split. This implementation uses a reduction to weighted one-against-all multi-class classification as employed in \cite{dudik2011}.
The hyper-parameters are the same as in the Regression approach.
\item \textbf{DRO} \cite{dudik2011}: Combines the Regression method with IPS using the doubly robust estimator to perform policy optimization. Again uses a reduction to weighted one-against-all multi-class classification, and uses the same set of hyper-parameters.
\item \textbf{POEM} \cite{swaminathan2015batch}: Directly trains a stochastic policy following the counterfactual risk minimization principle, thus reasoning about differences in the variance of the IPS estimate $\hat{R}(\pi)$. Hyper-parameters are variance regularization, $L2$ regularization, propensity clipping and number of training epochs.
\end{enumerate}

\begin{table}
\centering
    \begin{tabular}{|l|ccc|}
    \toprule
     & \multicolumn{3}{|c|}{Test set estimates} \\
    Approach & $\hat{R}(\pi_\epsilon) \times 10^4$ & $\hat{R}(\pi_\epsilon) \times 10^4 / \hat{C}(\pi_\epsilon)$ & $\hat{C}(\pi_\epsilon)$ \\
    \midrule
Random & $44.676 \! \pm \! 2.112$ & $45.446 \! \pm \! 0.001$ & $0.983 \! \pm \! 0.021$ \\
$\pi_0$ & $53.540 \! \pm \! 0.224$ & $53.540 \! \pm \! 0.000$ & $1.000 \! \pm \! 0.000$ \\
Regression & $48.353 \! \pm \! 3.253$ & $48.162 \! \pm \! 0.001$ & $1.004 \! \pm \! 0.041$ \\
IPS & $54.125 \! \pm \! 2.517$ & $53.672 \! \pm \! 0.001$ & $1.008 \! \pm \! 0.016$ \\
DRO & $57.356 \! \pm \! 14.008$ & $57.086 \! \pm \! 0.005$ & $1.005 \! \pm \! 0.025$ \\
POEM & $58.040 \! \pm \! 3.407$ & $57.480 \! \pm \! 0.001$ & $1.010 \! \pm \! 0.018$ \\
    \bottomrule
    \end{tabular}
    \caption{Test set performance of policies learnt using different counterfactual learning baselines. Errors bars are $99\%$ confidence intervals under a normal distribution. Confidence interval for SNIPS is constructed using the delta method \cite{Owen2013}.}
    \label{tab:learnres}
\end{table}

The results of the learning experiments are summarized in Table~\ref{tab:learnres}. For more details and the specifics of the experiment setup, visit the dataset \href{http://www.cs.cornell.edu/~adith/Criteo/}{website}.
Differences in Random and $\pi_0$ numbers compared to Table~\ref{tab:epsilonlogger1} are because they are computed on a $33\%$ subset --- we do expect their confidence intervals to overlap.
We see that the \emph{Regression} approach, which loosely corresponds to predicting CTR for each candidate using supervised machine learning, can be substantially improved using many recent off-policy learning algorithms that effectively use the logged propensities.
We also note that very limited hyper-parameter tuning was performed for methods like \emph{POEM} and \emph{DRO} --- for instance, POEM can conceivably be improved by employing the doubly robust estimator. We leave such algorithm-tuning to future work.

\section{Conclusions}

In this paper, we have introduced a standardized test-bed to systematically investigate off-policy learning algorithms using real-word data. We presented this test-bed, the sanity checks we ran to ensure its validity, and showed results comparing 
state-of-the-art off-policy learning methods
(doubly robust optimization \cite{dudik2011} and
POEM \cite{swaminathan2015batch}) to regression baselines on a $1$-slot banner filling task.
Our results show experimental evidence that recent off-policy learning methods can improve upon state-of-the-art supervised learning techniques on a large-scale real-world data set.

These results we presented are for the 1-slot banner filling tasks. There are several dimensions in setting up challenging, interesting, relevant off-policy learning problems on the data collected for future work.

\begin{description}
\item[Size of the action space:] Increase the size of the action space, i.e. of the number of slots in the banner. 
\item[Feedback granularity:] We can use global feedback (was there a click somewhere in the banner), or per item feedback (which item in the banner was clicked).
\item[Contextualization:] 
We can learn a separate model for each banner type or learn a contextualized model across multiple banner types.
\end{description}

\subsubsection*{Acknowledgments}
We thank Alexandre Gilotte and Thomas Nedelec at Criteo for their help in creating the dataset.
This work was funded in part through NSF Awards IIS-1247637, IIS-1615706, IIS-1513692. 



\bibliography{refs}
\bibliographystyle{ieeetr}


\end{document}